\documentclass{IEEEtran}
\usepackage{amsfonts}
\usepackage{amsgen,amsmath,amstext,amsbsy,amsopn,amssymb,amsthm}
\usepackage{times}
\usepackage{graphicx} % more modern
\usepackage{subfigure} 

\usepackage{algorithm}
\usepackage{algorithmic}
\usepackage{multirow}
\usepackage{hyperref}
\usepackage{caption}

\newcommand{\argmax}{\mathop{\rm arg\max}}

\newcommand{\bbR}{\mathbb{R}}

\newcommand{\cR}{{\mathcal{R}}}

\newcommand{\0}{{\mathbf{0}}}

\renewcommand{\t}{{\mathbf{t}}}

\newcommand{\x}{{\mathbf{x}}}
\newcommand{\y}{{\mathbf{y}}}

\newcommand{\A}{{\mathbf{A}}}
\newcommand{\B}{{\mathbf{B}}}
\newcommand{\C}{{\mathbf{C}}}
\newcommand{\D}{{\mathbf{D}}}
\newcommand{\I}{{\mathbf{I}}}

\newcommand{\U}{{\mathbf{U}}}
\newcommand{\w}{{\mathbf{w}}}
\newcommand{\V}{{\rm V}}

\newcommand{\bPhi}{\mathbf{\Phi}}
\newcommand{\bphi}{\boldsymbol{\phi}}

\newcommand{\bmu}{\boldsymbol{\mu}}
\newcommand{\bSigma}{\boldsymbol{\Sigma}}
\newcommand{\bepsilon}{\boldsymbol{\epsilon}}
\newcommand{\btheta}{\boldsymbol{\theta}}

\newcommand{\bLambda}{\boldsymbol{\Lambda}}

\newcommand{\tr}{{\rm tr}}

\newtheorem{theorem}{Theorem}

\title{Bayesian linear regression with Student-t assumptions}
\author{Chaobing Song, Shu-Tao Xia}
\begin{document} 
\maketitle
% \twocolumn[
% \icmltitle{Bayesian linear regression with Student-t assumptions}

% It is OKAY to include author information, even for blind
% submissions: the style file will automatically remove it for you
% unless you've provided the [accepted] option to the icml2016
% package.
% \icmlauthor{Chao-Bing Song$^{*}$}{songchaobin@126.com}
% \icmladdress{Room H205E, Tsinghua Park, University Town of XiLi, NanShan District, ShenZhen City, GuangDong Province, China}
% \icmlauthor{Shu-Tao Xia$^{*}$}{xiast@sz.tsinghua.edu.cn}
% \icmladdress{$^{*}$Tsinghua Park, University Town of XiLi, NanShan District, ShenZhen City, GuangDong Province, China}

% You may provide any keywords that you 
% find helpful for describing your paper; these are used to populate 
% the "keywords" metadata in the PDF but will not be shown in the document
% \icmlkeywords{Gaussian process, Bayesian network, Student-t distribution}

% \vskip 0.3in

\begin{abstract} 
% Gaussian process have attracted a lot of attention in recent years. 
As an automatic method of determining model complexity using the training data alone, Bayesian linear regression provides us a principled way to select hyperparameters. 
But one often needs approximation inference if distribution assumption is beyond Gaussian distribution. In this paper, we propose a Bayesian linear regression model with Student-t assumptions (BLRS), which can be inferred exactly. 
 In this framework, 
 both conjugate prior and expectation maximization (EM) algorithm are generalized. Meanwhile, we prove that the maximum likelihood solution is equivalent to the standard Bayesian linear regression with Gaussian assumptions (BLRG). The $q$-EM algorithm for BLRS is nearly identical to the EM algorithm for BLRG. It is showed that $q$-EM for BLRS can converge faster than EM for BLRG for the task of predicting online news popularity.

\end{abstract} 
\begin{section}{Introduction}
In a practical application of linear regression, in order to achieve the best prediction performance on new data, one must find a model with appropriate complexity parameters that governs model complexity. One popular method to tune complexity parameters is cross validation \cite{shao1993linear}, which uses a proportion $(K-1)/K$ of the available data for training while making use of all the data to assess performance. However, the number of training runs that must be performed is proportional to $K$, and it is proven to be problematic if the training is computational expensive or $K$ must be set to a large value because of the scarce of data. Furthermore, if a model has several complexity parameters, in the worst case, searching the combinations of the complexity parameters needs a number of training runs that is exponential to the number of parameters.

As an alternative to cross validation, one can turn to a Bayesian treatment of linear regression, which introduces prior probability distribution of weight parameters and noise. Then after marginalizing the weight parameters out, the hyperparameters of the prior probability distributions can be estimated by maximizing the probability of observations. Thus in Bayesian linear regression, only training data is needed to tune hyperparameters which correspond to complexity parameters.  However, although a principled way to estimate hyperparameters is provided, but the established model from Bayesian perspective is often difficult to solve and in many cases stochastic or deterministic approximation techniques are often required. A rare exception is Bayesian linear regression with independent Gaussian prior distribution of weight parameters and noise (we abbreviate the model as ``BLRG''), in which the posterior distribution can be expressed in a closed form and then the model can be inferred exactly by expectation maximization (EM) algorithm. 

An interesting question is whether we can find more generalized models which still keeps a closed-form posterior distribution and therefore can be inferred exactly. In this paper, we give a positive answer. Observing the form of distribution, Gaussian distribution is a limit case of Student-t distribution for the degree of freedom $
\nu\rightarrow+\infty$. In addition, from the perspective of nonextensive information theory \cite{tsallis2001nonextensive},  
Student-t and Gaussian distribution both are the maximum Tsallis entropy distribution \cite{de1997student} of which the Gaussian distribution is a special case for entropy index $q=1$.  Based on these facts and the results from nonextensive information theory, we unify the inference process for the assumptions of Gaussian and Student-t distributions and propose a Bayesian linear regression model with Student-t assumptions (``BLRS''). The main contributions of this paper are
\begin{itemize}
	\item  By introducing relevant noise whose variance is linear to the norm of weight parameters, we generalize the concept of conjugate prior to Student-t distribution, e.g, under the relevant noise setting, if the prior distribution of weight parameters is Student-t distribution, the posterior distribution is also a Student-t distribution.
	\item We prove that the maximum likelihood solution of variance hyperparameters has no relation with the degrees of freedom $\nu$ and thus BLRS is equivalent to BLRG, which may be remarkable. 
	% Therefore, $\nu$ in the proposed model can be seen as an accelerating factor for iteration.
	\item By applying Tsallis divergence instead of Kullback-Leibler (KL) divergence, EM is generalized  to $q$-EM to make an exact inference under the setting of Student-t distribution. Closed-form expressions are acquired in each iteration, which are nearly identical to the EM algorithm for BLRG.
	\item By experiments on the task of predicting online news popularity, we show that BLRS and BLRG will converge to the same result. Meanwhile, BLRS with a finite constant $\nu$ can converge faster to BLRG on standard datasets. Therefore, $\nu$ can be seen as an accelerating factor for iterations. 
	% \item By maximizing the probability of observations (model evidence) with respect to the degree of freedom $\nu$, $\nu$ is determined automatically. A extreme case is fixing $\nu$ to $+\infty$. For $\nu\rightarrow+\infty$, the model is degraded to the standard Bayesian linear regression with Gaussian assumptions. 
\end{itemize}

In addition, when preparing this paper, we find that in terms of the general form, the $\alpha$-EM algorithm in \cite{matsuyama2003alpha} is equivalent to the $q$-EM algorithm we proposed in this paper. But they still have $3$ differences.  Firstly, $q$-EM is mainly specified to solve the corresponding BLRS model, while $\alpha$-EM attempts to improve EM without changing model. Secondly, $q$-EM is derived from nonextensive information theory, while $\alpha$-EM is derived from information geometry. Thirdly, the $q$-EM algorithm is one part of our attempt to unify the treatment to Gaussian distribution and Student-t distribution, while $\alpha$-EM and its extensions are with independent interests \cite{matsuyama2003alpha}. 

% the value $q$ is specialized to the degree of freedom, while in \cite{matsuyama2003alpha} $\alpha$ is used by searching in a particular range. Secondly, Matsuyama's goal is to improve the speed of EM algorithm and emphasize the efficiency, while $q$-EM is proposed to formulate an exact inference algorithm and emphasize the effectiveness. 

% we find that the $\alpha$-EM algorithm in \cite{matsuyama2003alpha} is equivalent to the $q$-EM algorithm we proposed in this paper. The author's goal is to improve the speed of EM algorithm and emphasize the efficiency, while $q$-EM is proposed to formulate a solvable exact inference algorithm and emphasize the effectiveness. In addition, we get inspired from the nonextensive information theory and emphasize its application in the concrete Bayesian linear regression, while his goal is to formulate a general framework and use the term from information geometry. In addition, the value $q$ is specialized to the degree of freedom, while in \cite{matsuyama2003alpha} $\alpha$ is used by search in particular range.  Finally, we $q$-EM algorithm is only one part of our attempt to unify the treatment to Gaussian distribution and Student-t distribution, while $\alpha$-EM and its extensions are the main body of \cite{matsuyama2003alpha}.

% we consider $q$ (corresponding to the degree of freedom) or $\alpha$

% Gaussian process have attracted a lot of attentions in recent years. As a non-parametric kernel machine,  it combines the advantages of Bayesian network and kernel machine. In the view of Gaussian process, feature map and prior. 
\end{section}

\begin{section}{Notations and concepts from nonextensive information theory}
Nonextensive information theory is proposed by Tsallis \cite{tsallis1988possible} and aims to model phenomenon such as long-range interactions and multifractals \cite{tsallis2001nonextensive}. Nonextensive information theory has recently been applied in machine learning \cite{martins2009nonextensive}, \cite{ding2010t}. In this paper, it is our main motivation for generalizing BLRG. In this section, we review some notations and concepts briefly.

For convenience \cite{tsallis2001nonextensive}, one can define the following $q$-exponent and $q$-logarithm,
\begin{eqnarray*}
\exp_{q}x&=&
\begin{cases}
[1+(1-q)x]_+^{\frac{1}{1-q}} & q\in\bbR\backslash\{1\}\\
\exp x    & q=1 \\
\end{cases},  \\
\ln_{q}x&=&\begin{cases}
	\frac{x^{1-q}-1}{1-q} &  q\in\bbR\backslash\{1\}\\
	\ln x & q=1
\end{cases}.
\end{eqnarray*}
where $[x]_+$ stands for $\max\{x,0\}$ and $\exp_1 x=\lim_{q\rightarrow 1}\exp_q x$, $\ln_1x=\lim_{q\rightarrow1}\ln_q x$.
 By its definition, one has
\begin{eqnarray*}
\exp_{q}(\ln_{q}x)&=&x,\\
\ln_{q}(\exp_{q}x)&=&x.
\end{eqnarray*}
The $q$-exponent and $q$-logarithm has the following properties,
\begin{eqnarray}
\exp_q(xy)&=&\exp\left(\frac{1}{1+(1-q)y} x\right)\exp(y),\label{eq:expq}\\
\ln_q(xy)&=&\ln_q(x)+\ln_q(y)\nonumber\\
&& +(1-q)\ln_q(x)\ln_q(y),\label{eq:logq}
\end{eqnarray}
where \eqref{eq:expq} is used to formulate a generalized conjugate prior for Student-t distribution and \eqref{eq:logq} is used to generalize the EM algorithm.

For  normalized probability distributions $p(\x)$ and $t(\x)$ on  $\x\in R^n$, from nonextensive information theory, Tsallis divergence \cite{cichocki2010families} is defined as 
% \[
% D_q(p)=\frac{num}{den}
% ]
\begin{eqnarray*}
D_{q}(p(\x)\|t(\x))
&=&\int -p(\x)\ln_q\left(\frac{t(\x)}{p(\x)}\right)d\x\\
&=&
\begin{cases}
\frac{\int p^{q}(\x)t^{1-q}(\x)d\x-1}{q-1}, & q\in\cR\backslash\{1\} \\
\int p(\x)\ln\frac{p(\x)}{t(\x)}d\x, & q = 1
\end{cases},
\end{eqnarray*}
At $q=1$, denote $KL(p(\x)\|t(\x))=D_1(p(\x)\|t(\x))=\lim_{q\rightarrow1}D_q(p(\x)\|t(\x))$,
which is the definition of KL divergence.

For $q>0$, $D_q(p(\x)\|t(\x))$ is a special case of f-divergence (see \cite{cichocki2010families} and reference therein), which has the following properties:
\begin{itemize}
\item Convexity: $D_q(p(\x)\|t(\x))$ is convex with respect to both $p(\x)$ and $t(\x)$;
\item Strict Positivity: $D_q(p(\x)\|t(\x))\ge0$ and $D_q(p(\x)\|t(\x))=0$ if and only if $p(\x)=t(\x)$.
\end{itemize}
Because of the two useful properties for $q>0$, the value of $D_q(p(\x)\|t(\x))$ can be used to measure the similarity between $p(\x)$ and $t(\x)$. In practice,
one can make $p(\x)$ get close to $t(\x)$ as much as possible by minimizing $D_q(p(\x)\|t(\x))$ with respect to $p(\x)$. 
In the $q$-EM algorithm, Tsallis divergence is used to measure similarity and alleviate the complexity of mathematical expressions. 
\end{section}

\begin{section}{Bayesian linear regression with Gaussian assumptions}\label{sec:lr}
Given a set of  pairs $(\x_i, y_i),$ $i\in\{1,2,\ldots,m\}$ $,\x_i\in\bbR^n,y_i\in\bbR$. $\{\phi_i(\cdot)\}_{i=1}^M$ is a group of fixed basis functions.  Consider the linear model
\begin{eqnarray*}
\y=\bPhi \w+\bepsilon,
\end{eqnarray*}
where $\w\in\bbR^{M}$ is the parameter we need to estimate, $\bepsilon$ is an additive noise, $\bPhi=(\bphi_1,\bphi_2,\ldots,\bphi_M)^T$ with $\bphi_i=(\phi_i(\x_1),\phi_i(\x_2),\ldots,\phi_i(\x_m))^T, i=1,2,\ldots,M$.
In Bayesian linear regression with Gaussian assumptions (BLRG), $\bepsilon\in\bbR^m$ is assumed to be an independent, zero-mean Gaussian noise,
\begin{equation}
 p(\bepsilon;\beta)= N(0,\beta^{-1}\I),
 \end{equation} 
where $\beta$ is the inverse variance.
 % and $\boldsymbol{\epsilon}=(\epsilon_1,\epsilon_2,\cdots,\epsilon_m)^T$.
Meanwhile, $\w$ is assumed to be an independent zero-mean Gaussian distributed random variable, given by
\begin{eqnarray*}
p(\w;\alpha)= N(0,\alpha^{-1}\I),
\end{eqnarray*}
where $\alpha$ is the inverse variance of $\w$. 
Then the likelihood is 
\[
p(\y|\w;\alpha,\beta)=N(\y|\bPhi\w,\beta^{-1}\mathbf{I}).
\]

In this setting, by integrating out $\w$ in $p(\y,\w;\alpha,\beta)=p(\y|\w;\alpha,\beta)p(\w;\alpha)$ and applying Bayesian theorem, one has
\begin{eqnarray*}
p(\y;\alpha,\beta)&=& N(0,\beta^{-1}\I+\alpha^{-1}\bPhi\bPhi^T),\\
p(\w|\y;\alpha,\beta)&=& N(\boldsymbol{\bmu},\bSigma),
\end{eqnarray*}
where
\begin{eqnarray*}
	\bSigma&=&(\beta\bPhi^T\bPhi+\alpha \I)^{-1},\\
	\bmu&=&\beta\bSigma\bPhi^T \t=(\bPhi^T\bPhi+\frac{\alpha}{\beta} \I)^{-1}\bPhi^T\y. 
\end{eqnarray*}
Fixed $\alpha,\beta$, $\bmu$ is equivalent to the solution of ridge regression \cite{hoerl1970ridge} with regularization parameter $\frac{\alpha}{\beta}$. The hyperparamters $\alpha,\beta$ can be optimized by maximum likelihood principle, e.g., maximizing $\ln p(\y;\alpha,\beta)$ with respect to $\alpha, \beta$, 
\begin{eqnarray*}
(\alpha,\beta)
&=&\argmax_{\alpha,\beta}\Big\{-\ln|\beta^{-1}\I+\alpha^{-1}\bPhi\bPhi^T| \\
&& -\y^T(\beta^{-1}\I+\alpha^{-1}\bPhi\bPhi^T)^{-1}\y\Big\}.
\end{eqnarray*}
Gradient-based optimization method such as conjugate conditional method or quasi Newton method with active set strategy can be used to solve the above problem under the nonnegative constraints $\alpha>0,\beta>0$. However, there exist an elegant and powerful algorithm called expectation maximization (EM) algorithm to address this problem. \cite{redner1984mixture} concluded that the EM algorithm have the advantage of reliable global convergence, low cost per iteration, economy of storage and ease pf programming. Consider a general joint distribution $p(\y,\w; \btheta)$, where $\y$ is  observations, $\w$ is hidden variables and $\btheta$ is parameters needed to optimize. In order to optimize the evidence distribution $p(\y;\btheta)$, the general EM algorithm is executed in Alg. \ref{alg:em}.

\begin{algorithm}[htb]
   \caption{Expectation maximization}
   
% \begin{algorithmic}
\begin{enumerate}
\item Choose an initial setting for the parameters $\btheta^{\text{old}}$;
\item E Step: Evaluate $p(\w|\y;\btheta^{\text{old}})$;
\item M Step: Evaluate $\btheta$, given by
\[
\btheta =\argmax_{\btheta} -KL((p(\w|\y;\btheta^{\text{old}})\|p(\y,\w;\btheta));
\] 
\item Check for convergence of either the log likelihood or the parameter values. If the convergence criterion is not satisfied, then let
\[
\btheta^{\text{old}}=\btheta
\]
and return to step 2.
\end{enumerate}
% \end{algorithmic}
\label{alg:em}
\end{algorithm}

The concrete process of EM for BLRG is a special case of the $q$-EM iteration for BLRS and will be discussed in Section \ref{sec:q-EM}.

\end{section}

\begin{section}{Generalized conjugate prior for Student-t distribution}
\label{sec:conjugate}
Conjugate prior is an important concept in Bayesian inference. If the posterior distributions $p(\w|\y;
\btheta)$ are in the same family as the prior probability distribution $p(\w;\btheta)$, the prior and posterior are then called conjugate distributions, and the prior is called a conjugate prior for the likelihood function. Conjugate prior is often thought as a particular characteristic of exponential family \cite{diaconis1979conjugate}.  
From the view of mathematics, it is because Bayesian update is multiplicative. If the  prior distribution or the likelihood distribution  is not belonged to exponential family, the product of the two distributions, e.g, joint distribution will exist cross term in general, which makes the integral over $\w$ be intractable. To alleviate this complexity, in Bayesian linear regression model, we assume that not only the expectation of the likelihood $p(\y|\w;\alpha,\beta)$, but also the variance dependent on the weight parameters $\w$.

Firstly,
 % converse with the hierarchical prior in Section \ref{sec:lr} which implies only marginal Student-t distribution,
  we assume that $\w$ is distributed as joint Student-t distribution
\begin{eqnarray}
&&p(\w;\nu,\alpha)=\text{St}(\w|\nu,\mathbf{0},\alpha^{-1}\I)\nonumber\\
&=&\frac{\Gamma\left((\nu+M)/2\right)\alpha^{M/2}}{\Gamma(\nu/2)(\pi\nu)^{M/2}}\left(1+\frac{\alpha}{\nu}\|\w\|_2^2\right)^{-\frac{\nu+M}{2}}. \label{eq:w-st}
\end{eqnarray}
Then we assume that noise is distributed as 
\begin{eqnarray*}
 &&p(\bepsilon;\nu, \alpha,\beta)\\
 &=&\text{St}\left(\bepsilon|\nu+M, \0, \frac{\nu}{\nu+M}\left(1+\frac{\alpha}{\nu}\|\w\|_2^2\right)\beta^{-1}\I\right),
\end{eqnarray*}  
where the degree of freedom of $\bepsilon$ is $M$ greater than that of $\w$ and $\w$ influences the variance of $\bepsilon$ by a product factor $1+\frac{\alpha}{\nu}\|\w\|_2^2$.
Then the likelihood 
\begin{eqnarray}
&&p(\y|\w;\nu, \alpha,\beta)\nonumber\\
&=&\text{St}(\y|\nu+M, \bPhi\w, \frac{\nu}{\nu+M}\left(1+\frac{\alpha}{\nu}\|\w\|_2^2\right)\beta^{-1}\I)\nonumber\\
&=&\frac{\Gamma(\frac{\nu+M+m}{2})\beta^{\frac{m}{2}}}{\Gamma(\frac{\nu+M}{2})\pi^{\frac{m}{2}}\left(\nu+\alpha\|\w\|_2^2\right)^{\frac{m}{2}}}\nonumber\\
&&\!\!\!\!\!\!\!\!\!\!\cdot\left(1+\frac{\beta}{\nu+\alpha\|\w\|_2^2}\|\y-\Phi\w\|_2^2\right)^{-\frac{\nu+M+m}{2}} \label{eq:yw-st}
\end{eqnarray}
Multiply \eqref{eq:w-st} by \eqref{eq:yw-st}, the joint distribution is 
\begin{eqnarray}
&&p(\y,\w;\nu, \alpha,\beta)\nonumber\\
&=&\frac{\Gamma(\frac{\nu+M+m}{2})\alpha^{\frac{M}{2}}\beta^{\frac{m}{2}}}{\Gamma(\frac{\nu}{2})(\nu\pi)^{(m+M)/2}}\nonumber\\
&&\!\!\!\!\!\!\!\!\!\!\cdot\left(1+\frac{\alpha}{\nu}\|\w\|_2^2+\frac{\beta}{\nu}\|\y-\Phi\w\|_2^2\right)^{-\frac{\nu+M+m}{2}}. \label{eq:joint}
\end{eqnarray}
% Denote  $\alpha^{\prime}=\frac{\alpha}{\nu}, \beta^{\prime}=\frac{\beta}{\nu}$, then
% \eqref{eq:joint} can be rewritten as
% \begin{eqnarray}
% &&p(\y,\w;\nu,\alpha^{\prime},\beta^{\prime})\nonumber\\
% &=&\frac{\Gamma(\frac{\nu+M+m}{2})(\alpha^{\prime})^{\frac{M}{2}}(\beta^{\prime})^{\frac{m}{2}}}{\Gamma(\frac{\nu}{2})\pi^{(M+m)/2}}\nonumber\\
% &&\cdot\left(1+\alpha^{\prime}\|\w\|_2^2+\beta^{\prime}\|\y-\Phi\w\|_2^2\right)^{-\frac{\nu+M+m}{2}}.\label{eq:joint2}
% \end{eqnarray}
% and
In addition,
\begin{eqnarray*}
&&\alpha\|\w\|_2^2 + \beta\|\y-\Phi\w\|_2^2\\
&=&\!\!\!\!\!(\w-\bmu)^T \A^{-1} (\w-\bmu)+\y^T\B^{-1} \y
% &&=(\w-\w_m)^T A (\w-\w_m)+\beta^{\prime}\t^T(\t-\Phi\w_m)\\
% &&=(\w-\w_m)^T A (\w-\w_m)+\|\t-\Phi\w_m\|_2^2+\alpha^{\prime}\w_m^T\w_m,
\end{eqnarray*}
where
\begin{eqnarray}
\A&=&\left(\alpha\I+\beta\bPhi^{T}\bPhi\right)^{-1},\label{eq:A}\\
\bmu&=&\beta \A\bPhi^{T}\y=(\bPhi^T\bPhi+\frac{\alpha}{\beta} \I)^{-1}\bPhi^T\y.,\label{eq:bmu}\\
\B&=&\beta^{-1}\I+\alpha^{-1}\bPhi\bPhi^T.\label{eq:B}
\end{eqnarray}
Then
\begin{eqnarray}
&&p(\y;\nu,\alpha,\beta)\nonumber\\
&=&\frac{\Gamma(\frac{\nu+m}{2})}{\Gamma(\frac{\nu}{2})(\nu\pi)^{\frac{m}{2}}|\B|^{1/2}}
 \left(1+\frac{1}{\nu}\y^T\B^{-1}\y\right)^{-\frac{\nu+m}{2}}\nonumber\\
&=&\text{St}(\y|\nu,\0,\B) \label{eq:observ}
\end{eqnarray}
and
\begin{eqnarray}
&&p(\w|\y;\nu, \alpha,\beta) \nonumber\\
&=&\frac{\Gamma(\frac{\nu+M+m}{2})}{\Gamma(\frac{\nu+m}{2})\pi^{\frac{M}{2}}(\nu+\y^T\B^{-1}\y)^{\frac{M}{2}}|\A|^{1/2}}\nonumber\\
&&\cdot\left(1+\frac{(\w-\bmu)^T \A^{-1}(\w-\bmu)}{\nu+\y^T\B^{-1}\y}\right)^{-\frac{\nu+m+M}{2}}\nonumber\\
&=& \text{St}(\w|\nu+m,\bmu,\C),\label{eq:posterior}
\end{eqnarray}
where
\begin{eqnarray}
\C&=&\frac{1}{\nu+m}(\nu+\y^T\B^{-1}\y)\A,\label{eq:C}
\end{eqnarray}
and $\bmu$ is given in \eqref{eq:bmu}.

Therefore, the posterior distribution $p(\w|\y;\nu,\alpha,\beta)$ is also a Student-t distribution with the increased degree of freedom $\nu+m$, which represents the property of  \emph{conjugate prior}. Just like the case in Gaussian assumptions, all the distributions related above have closed forms. If $\nu\rightarrow+\infty$, all the distributions will degraded to Gaussian distribution and thus the BLRG model is recovered.

In the above derivation, the product factor $1+\frac{\alpha}{\nu}\|\w\|_2^2$  plays a key role for the closed-form distributions, which is the main difference from common model settings about Student-t assumptions \cite{tzikas2008variational}. We propose it according to \eqref{eq:expq}, which is a generalization of the property of exponent. Combing the relevant noise, the property of conjugate prior is generalized naturally to Student-t distribution. Of course the thought of relevant noise can also be used to other kinds of distribution which is not belonged to exponential family, such as generalized Pareto distribution, which is beyond the scope of the paper. 
% Although it is proposed mainly to alleviate the computational complexity, it also has some physical meaning.
% It assumes that variance of noise will increase as the norm (or energy) $\|\w\|_2^2$ increase, which implies that the noise can not be neglected even if the energy of signal is 
% \begin{eqnarray}
% &&p(\y|\w;\alpha,\beta)\nonumber\\
% &=&\text{St}(\y|\nu+m, \bPhi\w, (1+\|\w\|_2^2)\beta^{-1}\I)\nonumber\\

% \end{eqnarray}

% \begin{eqnarray*}
% p(\y,\w;\alpha,\beta)=\frac{\Gamma\left(\frac{\nu+M}{2}\right)}{\Gamma\left(\frac{\nu}{2})}
% \end{eqnarray*}
\end{section}

\section{Maximum likelihood solution}
In this section, we consider the determination of the model parameters of BLRS using maximum likelihood. 
According to type-II maximum likelihood principle, one need to maximize \eqref{eq:observ} with respect to $\nu,\alpha,\beta$. Firstly,  we assume that the eigenvalues decomposition of $\bPhi\bPhi^T$ is $\bPhi\bPhi^T=\U\bLambda\U^T$, where $\U\in\bbR^{m\times m}$ is an orthogonal matrix, $\bLambda$ is a diagonal matrix with all diagonal elements $\lambda_1\ge\lambda_2\ge\cdots\lambda_m\ge 0$. Then 
\begin{eqnarray*}
\y^T\B^{-1}\y&=&\y^T(\beta^{-1}\I+\alpha^{-1}\U\bLambda\U^T)^{-1}\y\\
&=&(\U^T\y)^T(\beta^{-1}\I+\alpha^{-1}\bLambda)^{-1}(\U^T\y),
\end{eqnarray*}
Denote $\bar{\y}=\U^T\y=(\bar{y}_1,\bar{y}_2,\ldots,\bar{y}_m)^T$. 
Then
\begin{eqnarray*}
\y^T\B^{-1}\y&=&\sum_{i=1}^{m}\bar{y}_i^2(\beta^{-1}+\alpha^{-1}\lambda_i)^{-1}\\
&=&\beta\sum_{i=1}^{m}\bar{y}_i^2(1+\lambda_i\frac{\beta}{\alpha})^{-1}
\end{eqnarray*}
In addition,
\begin{eqnarray*}
|\B|&=&\Pi_{i=1}^{m}(\beta^{-1}+\lambda_i\alpha^{-1})\\
&=&\beta^{-m}\Pi_{i=1}^{m}\left(1+\lambda_i\frac{\beta}{\alpha}\right)
\end{eqnarray*}
Consider $\nu$ is fixed. $\beta, \gamma=\frac{\beta}{\alpha}$ are two independent hyperparameters to optimize. Maximizing \eqref{eq:observ} with respect to $\beta,\gamma$ is equivalent to minimizing 
\begin{eqnarray}
f(\beta,\gamma)
&=&\left(|\B|^{-1/2}\left(1+\frac{1}{\nu}\y^T\B^{-1}\y\right)^{-\frac{\nu+m}{2}}\right)^{-\frac{2}{\nu+m}}\nonumber\\
&=&\beta^{-\frac{m}{\nu+m}}\Pi_{i=1}^{m}\left(1+\lambda_i\gamma\right)^{\frac{1}{\nu+m}}\nonumber\\
&&\cdot\Big(1
+\frac{\beta}{\nu}\sum_{i=1}^{m}\bar{y}_i^2(1+\lambda_i\gamma)^{-1}\Big)\label{eq:f-b-g}
\end{eqnarray}
Set $\frac{\partial f}{\partial \beta}=0$, one gets
\begin{eqnarray}
\beta=\frac{m}{\sum_{i=1}^{m}\bar{y}_i^2(1+\lambda_i\gamma)^{-1}},\label{eq:beta}
\end{eqnarray}
where $\beta$ is determined by data and $\gamma$.
Substitute \eqref{eq:beta} into \eqref{eq:f-b-g}, we have
\begin{eqnarray}
f(\gamma)&=&\left(\frac{\sum_{i=1}^{m}\bar{y}_i^2(1+\lambda_i\gamma)^{-1}}{m}\right)^{\frac{m}{\nu+m}}\nonumber\\
&&\cdot\Pi_{i=1}^{m}\left(1+\lambda_i\gamma\right)^{\frac{1}{\nu+m}}\left(1+\frac{m}{\nu}\right),\label{eq:f-g}
\end{eqnarray}
For fixed $\nu$, minimizing $f(\gamma)$ is equivalent to solve
\begin{eqnarray}
\min_{\gamma}\quad m\ln\left(\sum_{i=1}^{m}\bar{y}_i^2(1+\lambda_i\gamma)^{-1}\right)+\sum_{i=1}^m\ln(1+\lambda_i\gamma),\label{eq:gamma}
\end{eqnarray}
where $\gamma$ is determined by data. Combing \eqref{eq:beta} and \eqref{eq:gamma}, it is showed that whatever $\nu$ is, the maximum likelihood solution of $\beta,\gamma$ (i.e., $\alpha,\beta$) will be the same. For $\nu\rightarrow+\infty$, by maximizing $\ln p(\y;\alpha,\beta)$, \eqref{eq:beta} and \eqref{eq:gamma} will also be acquired. 
Therefore we have Theorem \ref{thm:1}.
% \begin{itemize}
% 	\item 同一组超参数（alpha,beta)可能对应着无数多个等价的概率模型，即使超参数通过BLRG有效的估计了，也不能说数据一定是独立的以及噪声和参数先验一定服从高斯分布；alpha, beta唯一决定了参数的均值，但对应的分布形式和协方差不唯一
% 	\item 在实际的点估计预测中，只有参数的均值被使用，而均值由alpha，beta唯一决定；因此在优化alpha，beta时可以根据需要选择任意大于1的nu进行
% 	\item 该结果有效的解释了为什么学生t模型不那么如预期有效的原因，因为其在特定的模型设置下与高斯模型等价
% \end{itemize}

\begin{theorem}\label{thm:1}
For $\nu\in(0, +\infty]$,
the maximum likelihood solution of $\alpha,\beta$ of maximizing the evidence distribution $p(\y;\alpha,\beta)$ in \eqref{eq:observ} is only determined by data $\{(\x_i, y_i)\}_{i=1}^{m}$ and  is not influenced by the degrees of freedom $\nu$.
\end{theorem}

It is generally accepted that Student-t distribution is a generalization of Gaussian distribution. Thus it is believed that the related conclusion about Student-t distribution is some kind of generalization of the corresponding result about Gaussian distribution. Concretely, it is to say that, the related result about Student-t distribution should depend on the distribution parameter $\nu$; meanwhile, for $\nu\rightarrow+\infty$, the result will degraded to the corresponding result about Gaussian distribution.
But Theorem \ref{thm:1} shows that the maximum likelihood solution of $\alpha,\beta$ has no relation with the distribution parameter $\nu$. It has the following meanings:
\begin{itemize}
\item One pair $(\alpha,\beta)$ of the maximum likelihood solution corresponds to the evidence distribution $p(\y;\alpha,\beta)$ with arbitrary $\nu>0$. Therefore, even if BLRG selects suitable hyperparameters $\alpha,\beta$  and thus the resulted parameter $\w$ fits model well, we still cannot assert that the observed data is generated from independent Gaussian distribution with independent Gaussian noise. 
\item Because BLRG  and BLRS are special cases of Gaussian process and t process respectively and has the same maximum likelihood solution of $\alpha,\beta$, Theorem \ref{thm:1} explains the conclusion in \cite{rasmussen2004gaussian} that `` t process is perhaps not as exciting as one might have hoped''.
\item
If the parameter $\w$ is only used for point estimation, $\nu$ can be fixed to an arbitrary positive value or $+\infty$ for computational consideration.
\end{itemize}

% This is a remarkable result, because all the result
\begin{section}{$q$-Expectation maximization}
\label{sec:q-EM}
\subsection{The general $q$-EM algorithm}
In the general EM algorithm Alg. \ref{alg:em}, the logarithm of $p(\y;\btheta)$ is splitted into two KL divergences. In the $q$-EM algorithm, we generalize the property to $q$-logarithm. From \eqref{eq:logq}, one has
\begin{eqnarray*}
% &&\ln_{q}(xy)\\
% &=&\ln_{q}x+\ln_q y+(1-q)\ln_q x \ln_q y\\
\ln_q x&=&\frac{1}{1+(1-q)\ln_q y}(\ln_q (xy)-\ln_q y)\\
&=&y^{q-1}(\ln_q (xy)-\ln_q y)
\end{eqnarray*}
Then combing $p(\y;\btheta)=\frac{p(\y,\w;\btheta)}{p(\w|\y;\btheta)}$ and introducing a variational distribution $s(\w)$, we have
\begin{eqnarray}
&&\ln_q p(\y;\btheta)\nonumber\\
&=&p^{q-1}(\w|\y;\btheta)(\ln_q p(\y,\w;\btheta)-\ln_q p(\w|\y;\btheta))\nonumber\\
&=&\frac{1}{\int s^q(\w) d\w}\int s^{q}(\w)p^{q-1}(\w|\y;\btheta)\nonumber\\
&&\cdot\big(\ln_q p(\y,\w;\btheta)-\ln_q p(\w|\y;\btheta)\big)d\w\nonumber\\
&=&\frac{1}{\int s^q(\w)d\w}\int \left(s^{\prime}(\w)\right)^q\big(\ln_q p(\y,\w;\btheta)\nonumber\\
&&-\ln_q p(\w|\y;\btheta)\big)d\w\nonumber\\
&=&\frac{(\int s^{\prime}(\w)d\w)^q}{\int s^q(\w)d\w}\int \left(\frac{s^{\prime}(\w)}{\int s^{\prime}(\w)d\w}\right)^q \nonumber\\
&&\cdot(\ln_q p(\y,\w;\btheta)-\ln_q p(\w|\y;\btheta))d\w\nonumber\\
&=&\frac{(\int s^{\prime}(\w)d\w)^q}{\int s^q(\w)d\w}\Big(-D_q(s^{\prime\prime}(
\w)||p(\y,\w;\btheta))\nonumber\\
&&+D_q(s^{\prime\prime}(\w)\|p(\w|\y;\btheta))\Big)\nonumber\\
&\ge&-\frac{(\int s^{\prime\prime}(\w)d\w)^q}{\int s^q(\w)d\w}D_q(s^{\prime\prime}(
\w)\|p(\y,\w;\btheta)),\label{eq:logq2}
\end{eqnarray}
where $s^{\prime}(\w)=s(\w)p^{\frac{q-1}{q}}(\w|\y;\btheta)$, $s^{\prime\prime}(\w)=\frac{s^{\prime}(\w)}{\int s^{\prime}(\w)d\w}$. For $q=1$, \eqref{eq:logq2} degrades to
\begin{eqnarray}
&&\ln p(\y;\btheta)\nonumber\\
&=&-\text{KL}(s(\w)\|p(\y,\w;\btheta))+\text{KL}(s(\w)\|p(\w|\y;\btheta))\nonumber\\
&\ge& -\text{KL}(s(\w)\|p(\y,\w;\btheta)).
\end{eqnarray}
Just like the standard EM-step \cite{bishop2006pattern}, \cite{tzikas2008variational}, but using $q$-logarithm and Tsallis divergence instead of logarithm and KL divergence, in E-Step, $\ln_q p(\y;\btheta)$ is optimized with respect to $s^{\prime\prime}(\w)$, given by
$$s^{\prime\prime}(\w)= p(\w|\y;\btheta^{\text{old}}).$$
Then in M-Step, $\ln_q p(\y;\btheta)$ is optimized with respect to $\btheta$, given by
\begin{eqnarray*}
\btheta =\argmax_{\btheta} -D_q(p(\w|\y;\btheta^{\text{old}})\|p(\y,\w;\btheta));	
\end{eqnarray*}
Corresponding to Alg. \ref{alg:em}, the $q$-EM algorithm is summarized in Alg. \ref{alg:q-em}.
% \begin{itemize}
% \item E-Step: 
% \item M-Step: 
% \begin{eqnarray*}
% \btheta^{NEW}&=&\arg\max_{\btheta^{NEW}}-D_q(s^{\prime\prime}(
% z)||p(\y,\w;\btheta^{NEW}))\\
% &=&\arg\min_{\btheta^{NEW}}D_q(p(\w|\y;\btheta^{OLD})||p(\y,\w;\btheta^{NEW}))\\
% \end{eqnarray*}
% \end{itemize}

\begin{algorithm}[h]
   \caption{$q$-Expectation maximization}
   
% \begin{algorithmic}
\begin{enumerate}
\item Choose an initial setting for the parameters $\btheta^{\text{old}}$;
\item E Step: Evaluate $p(\w|\y;\btheta^{\text{old}})$;
\item M Step: Evaluate $\btheta$, given by
\[
\btheta=\argmax_{\btheta} -D_q(p(\w|\y;\btheta^{\text{old}})\|p(\y,\w;\btheta));
\] 
\item Check for convergence of either the log likelihood or the parameter values. If the convergence criterion is not satisfied, then let
\[
\btheta^{\text{old}}=\btheta
\]
and return to step 2.
\end{enumerate}
\label{alg:q-em}
% \end{algorithmic}
\end{algorithm}
Compare Alg. \ref{alg:em} and Alg. \ref{alg:q-em}, it is showed that the only difference is that Tsallis divergence is used instead KL divergence in Step $3$. 

Similar to EM \cite{tzikas2008variational}, in each iteration, $p(\y;\theta)$ will not decrease. Therefore, $q$-EM is also a local minimizer. 
\subsection{Inference and optimization with Student-t assumptions}
% Denote
% \begin{eqnarray*}
%  \bar{\B}&=&\nu\B\\
%  &=&\bar{\beta}^{-1}\I+\bar{\alpha}^{-1}\bPhi\bPhi^T,
%  \end{eqnarray*} 
% where $\bar{\beta}=\frac{\beta}{\nu}$ and $\bar{\alpha}=\frac{\alpha}{\nu}$,
% and
% \[
% \A=\left(\alpha\I+\beta\bPhi^{T}\bPhi\right)^{-1},\quad \bmu=\beta \A\bPhi^{T}\y.
% \]
% In order to optimize $\nu$ conveniently,  in the following setting, we optimize three independent hyperparameters $\nu,\bar{\alpha},\bar{\beta}$ instead of $\nu,\alpha,\beta$. 
% Then \eqref{eq:observ} and \eqref{eq:posterior} can be written as
% \begin{eqnarray}
% p(\y;\nu,\bar{\alpha},\bar{\beta})
% &=&\frac{\Gamma(\frac{\nu+m}{2})}{\Gamma(\frac{\nu}{2})\pi^{\frac{m}{2}}|\bar{\B}|^{1/2}}
%  \left(1+\y^T\bar{\B}^{-1}\y\right)^{-\frac{\nu+m}{2}}, \nonumber
% \end{eqnarray}
% and
% \begin{eqnarray}
% &&p(\w|\y;\nu, \alpha,\beta) \nonumber\\
% &=&\frac{\Gamma(\frac{\nu+M+m}{2})}{\Gamma(\frac{\nu+m}{2})(\nu\pi)^{\frac{M}{2}}(1+\y^T\bar{\B}^{-1}\y)^{\frac{M}{2}}|\A|^{1/2}}\nonumber\\
% &&\cdot\left(1+\frac{(\w-\bmu)^T \A^{-1}(\w-\bmu)}{\nu(1+\y^T\bar{\B}^{-1}\y)}\right)^{-\frac{v+m+M}{2}}\nonumber\\
% &=& \text{St}(\w|\nu+m,\bmu,\C),\label{eq:posterior-bar}
% \end{eqnarray}
% where
% \begin{eqnarray*}
% \bmu&=&\beta\A\bPhi^{T}\y,\\
% \C&=&\frac{\nu}{\nu+m}(1+\y^T\bar{\B}^{-1}\y)\A.
% \end{eqnarray*}
In this paper, we mainly use the mean $\bmu$ of $\w$ for point estimation, which is determined only by $\alpha,\beta$. Therefore, by Theorem \ref{thm:1}, $\nu$ can be set to a fixed positive value or $+\infty$ for computational consideration. In the following derivation,  $\nu$ is set to a positive constant. The case $\nu\rightarrow+\infty$ can be acquired by taking limit.

% In the model of Section \ref{sec:conjugate}, the task is to maximize $p(\y;\nu,\alpha,\beta)$ in \eqref{eq:observ} with respect to the three independent hyperparameters $\nu,\alpha,\beta$. Firstly, we consider the case $\nu$ is fixed. Then we consider the problem of optimizing $\nu$.

% Consider $\nu$ is fixed.
Before iteration, in Step $1$, we should specify some initial values of $\alpha,\beta$ as $\alpha^{\text{old}},\beta^{\text{old}}$. 
Then we apply $q$-EM to optimize the hyperparameters $\alpha,\beta$. 
% Following the step $1$ in Alg. \ref{alg:q-em}, if it is in the first iteration, $\alpha,\beta$ are set as the initial values selected before iteration. If having iterated once, $\alpha,\beta$ are set as the values of the last iteration.

Then in Step $2$, by \eqref{eq:posterior}, $p(\w|\y;\nu,\alpha^{\text{old}},\beta^{\text{old}})$ is a Student-t distribution, thus evaluating $p(\w|\y;\nu,\alpha^{\text{old}},\beta^{\text{old}})$ is equivalent to evaluating its values of distribution parameters given in \eqref{eq:A}, \eqref{eq:bmu}, \eqref{eq:B} and \eqref{eq:C} with $\alpha=\alpha^{\text{old}},\beta=\beta^{\text{old}}$. 

% \begin{eqnarray*}
% \A^{\text{old}}&=&\left(\alpha^{\text{old}}\I+\beta^{\text{old}}\bPhi^{T}\bPhi\right)^{-1},\\
% \bmu^{\text{old}}&=&\beta^{\text{old}} \A\bPhi^{T}\y,\\
% \B^{\text{old}}&=&(\beta^{\text{old}})^{-1}\I+(\alpha^{\text{old}})^{-1}\bPhi\bPhi^T,\\
% \C^{\text{old}}&=&\frac{1}{\nu+m}(\nu+\y^T\B^{-1}\y)\A.
% \end{eqnarray*}

% \begin{eqnarray}
% \A&=&\left((\alpha)^{\text{old}}\I+(\beta)^{\text{old}}\bPhi^{T}\bPhi\right)^{-1}, \label{eq:A}\\
% \bmu&=&(\beta)^{\text{old}}\A^{-1}\bPhi^T\y,\label{eq:bmu}\\
% \bSigma&=&\frac{1}{\nu+m}(1+ \nonumber\\
% &&\!\!\!\!\!(\beta)^{\text{old}}\y^T(\I-(\beta)^{\text{old}}\bPhi \A^{-1}\bPhi^T)\y)\A^{-1}. \label{eq:bSigma}
% \end{eqnarray}
Then in Step $3$, set 
$$\frac{1}{q-1}=\frac{\nu+M+m}{2},$$
which means $q>1$ if $\nu$ is a positive constant. For $\nu\rightarrow+\infty$, we set $q=1$.
and denote
% \begin{eqnarray*}
% C&=&\int p^q(\w|\y;\nu,\alpha^{\text{old}}, \beta^{\text{old}})d\w\\
% &=&\left(\frac{\Gamma\left(\frac{\nu+m+M}{2}\right)}{\Gamma\left(\frac{nu+m}{2}\right)\pi^{\frac{M}{2}}(\nu+y)}\\
% &=&\frac{\nu+m}{\nu+M+m}\\
% \end{eqnarray*}
% and 
\begin{eqnarray*}
p^{\prime}(\w)&=&\frac{1}{C} p^q(\w|\y;\nu,\alpha^{\text{old}}, \beta^{\text{old}}),
\end{eqnarray*}
where $C=\int p^q(\w|\y;\nu,\alpha^{\text{old}}, \beta^{\text{old}})d\w$ is constant for given $\alpha^{\text{old}},\beta^{\text{old}}$. Then 
$p^{\prime}(\w)$ is a normalized Student-t distribution with the degree of freedom $\nu+m+2$. It is deserved to note that only if $\nu>0$, the mean and covariance of  $p^{\prime}(\w)$ are both exist.
In addition, 
\begin{eqnarray}
E_{p^{\prime}}(\w)&=&\bmu, \label{eq:epmu}\\
E_{p^{\prime}}((\w-\bmu)(\w-\bmu)^T)&=&\C, \label{eq:epsigma}\\ 
E_{p^{\prime}}(\w\w^T)&=&\bmu\bmu^T+\C. \label{eq:mumuT}
\end{eqnarray}
% and
% \begin{eqnarray*}
% &&E_{p^{\prime}}(\w\w^T)\\
% &=&E_{p^{\prime}}((\w-\bmu)(\w-\bmu)^T)+\bmu\bmu^T\\
% &=&\frac{1}{\nu+m}(1+(\beta)^{\text{old}}\y^T(\I-(\beta)^{\text{old}}\bPhi \A^{-1}\bPhi^T)\y)A^{-1}\\
% &&+\bmu\bmu^T.\\
% \end{eqnarray*}
In \eqref{eq:joint}, denote
\[
Z=\frac{\Gamma(\frac{\nu}{2})(\nu\pi)^{(M+m)/2}}{\Gamma(\frac{\nu+M+m}{2})}.
\]
By the definition of Tsallis divergence, for $q>1$, maximizing $-D_q(p(\w|\y;(\alpha)^{\text{old}}, (\beta)^{\text{old}})\|p(\y,\w;\alpha, \beta))$ with respect to $\alpha, \beta$ is equivalent to minimizing $F(\alpha, \beta)$ given by
\begin{eqnarray*}
&&F(\alpha,\beta)\\
&=&\int p^q\left(\w|\y;\nu,(\alpha)^{\text{old}}, (\beta\right)^{\text{old}})p^{1-q}(\y,\w|\nu,\alpha,\beta)d\w\\
&=&C\int p^{\prime}(\w)p^{1-q}(\t,\w|\nu, \alpha,\beta))d\w\\
%=&&C\int p^{\prime}(\w)p^{1-q}(\t,\w|\alpha,\beta,v_{1,{k+1}})d\w  \\
&=&\frac{C}{Z^{1-q}}\alpha^{-\frac{M}{\nu+M+m}}\beta^{-\frac{m}{\nu+M+m}}\\
&&\cdot\int p^{\prime}(\w)\Big(1+\frac{\alpha}{\nu}\|\w\|_2^2
+\frac{\beta}{\nu}\|\y-\Phi\w\|_2^2\Big)d\w\\
&=&\frac{C}{Z^{1-q}}\alpha^{-\frac{M}{\nu+M+m}}\beta^{-\frac{m}{\nu+M+m}}\Big(1+\frac{\alpha}{\nu}\text{tr}(E_{p^{\prime}}(\w\w^T))\\
&&+\frac{\beta}{\nu}\left(\|\y\|_2^2
-2\y^T\bPhi E_{p^{\prime}}(\w)
+\text{tr}(\bPhi^T\bPhi E_{p^{\prime}}(\w\w^T)\right)\Big)
% &=&C\int p^{\prime}(\w)\frac{1}{Z^{1-q}}\Big(1+\alpha
% (\|
% \w-\mu_{k}\|_2^2
% +2\langle\w-\mu_{k},\mu_{k}\rangle+\|\mu_k\|_2^2)\\
% \quad &&+\beta(\|\y-\Phi\mu_{k}
% \|_2^2-2\langle t-\Phi\mu_{k},\Phi(\w-\mu_{k}\rangle+\|\Phi(\w-\mu_{k})\|_2^2)\Big)d\w \\
% &=&C\frac{1}{Z^{1-q}}\left(1+\frac{\alpha}{v_{1}-m}(\tr(\Sigma_{k})+\|\mu_{k}\|_2^2)+\frac{\beta}{v_1}(\|\t-\Phi\mu_k\|_2^2+\tr(\Phi^{T}\Phi\Sigma_{k}))\right)\\
% &=&C\left(\frac{\Gamma(\frac{v_1+n}{2})}{\Gamma(\frac{v_1-m}{2})}\right)^{\frac{2}{v_1+n}}\cdot\frac{\alpha^{-\frac{m}{v_1+n}}\beta^{-\frac{n}{v_1+n}}}{((v_1-m)\pi)^{-\frac{m}{v_1+n}}(v_1\pi)^{-\frac{n}{v_1+n}}}\\
% &&\cdot\left(1+\frac{\alpha}{v_{1}-m}(\tr(\Sigma_{k})+\|\mu_{k}\|_2^2)+\frac{\beta}{v_1}(\|\t-\Phi\mu_k\|_2^2+\tr(\Phi^{T}\Phi\Sigma_{k}))\right)
\end{eqnarray*}
Denote 
\begin{eqnarray}
b&=&\text{tr}(E_{p^{\prime}}(\w\w^T))=\|\bmu\|_2^2+\tr(\C)>0,\label{eq:b}\\
c&=&\|\y\|_2^2
-2\y^T\bPhi E_{p^{\prime}}(\w)
+\text{tr}(\bPhi^T\bPhi E_{p^{\prime}}(\w\w^T))\nonumber\\
&=&\|\y-\bPhi \bmu\|_2^2
+\text{tr}(\bPhi^T\bPhi\C)\label{eq:c}\\
&>&\nonumber0
\end{eqnarray}
% \begin{algorithm}[htb]
%    \caption{Expectation maximization}
%    \label{alg:em}
% % \begin{algorithmic}
% \begin{enumerate}
% \item Choose an initial setting for the parameters $\btheta^{\text{old}}$,
% \item E Step: Evaluate $p(\w|\y,\btheta^{\text{old}})$, 
% \item M Step: Evaluate $\btheta^{\text{new}}$, given by
% \[
% \btheta^{\text{new}} =\argmax -KL(p(\w|\y,\btheta^{\text{old}})\|p(\y,\w,\btheta^{new}),
% \] 
% \item Check for convergence of either the log likelihood or the parameter values. If the convergence criterion is not satisfied, then let
% \[
% \btheta^{\text{old}}=\btheta^{\text{new}}
% \]
% and return to step 2.
% \end{enumerate}
% % \end{algorithmic}
% \end{algorithm}
\end{section}
Set $\frac{\partial F(\alpha,\beta)}{\partial \alpha}=0$ and $\frac{\partial F(\alpha,\beta)}{\partial \beta}=0$, one has
\begin{eqnarray*}
-\frac{M}{\nu+M+m}\left(1+b\frac{\alpha}{\nu}+c\frac{\beta}{\nu}\right)+b\frac{\alpha}{\nu}&=&0,\\
-\frac{m}{\nu+M+m}\left(1+b\frac{\alpha}{\nu}+c\frac{\beta}{\nu}\right)+c\frac{\beta}{\nu}&=&0.
\end{eqnarray*}

Therefore,  
\begin{equation*}      
\left[             
  \begin{array}{cc}   
    (\nu+m)b & -Mc \\ 
   -mb & (\nu+M)c \\
  \end{array}
\right]\left[\begin{array}{c}
\alpha\\
\beta
\end{array}\right]   =\nu\left[\begin{array}{c}
M\\
m
\end{array} \right].           
\end{equation*}
By the fact 
\[
\left|            
  \begin{array}{cc}   
    (\nu+m)b & -Mc \\ 
   -mb & (\nu+M)c \\
  \end{array}
\right|=(\nu^2+(M+m)\nu)bc>0 , 
\]
after arrangement, we have
\begin{eqnarray}      
\left[\begin{array}{c}
\alpha\\
\beta
\end{array}\right]   &=& \nu\left[             
  \begin{array}{cc}   
    (\nu+m)b & -Mc \\ 
   -mb & (\nu+M)c \\
  \end{array}
\right]^{-1} \left[\begin{array}{c}
M\\
m
\end{array} \right]\nonumber\\
 &=&  \left[             
  \begin{array}{cc}   
    \frac{1}{b} & 0 \\ 
   0 &  \frac{1}{c} \\
  \end{array}
\right] \left[\begin{array}{c}
M\\
m
\end{array} \right] \label{eq:alpha-beta}
\end{eqnarray}
In \eqref{eq:alpha-beta}, it is interesting to see that given $b,c$, 
the final update formulation of $\alpha,\beta$ is identical to the EM update for BLRG \cite{tzikas2008variational}. The only difference is in the E step, where the covariance $\C$ in \eqref{eq:epsigma} has a product factor $\frac{\nu+\y^T\B^{-1}\y}{\nu+m}$, which will tend to $1$ for $\nu\rightarrow+\infty$. Summarizing the above steps, 
% $\nu$ does not influence $\alpha,\beta$ directly, but influences them indirectly by influencing $b,c$. For $\nu\rightarrow +\infty$, the expressions of $b,c$ degrade to the case with Gaussian assumptions. 
% In practice, only mean $\bmu$ in \eqref{eq:bmu} is used for point estimation, where $\bmu$ is only influenced directly by the ratio $\frac{\alpha}{\beta}$. Meanwhile, from \eqref{eq:alpha-beta}, one can see that $\frac{\alpha}{\beta}$ is influenced by $\nu$. Therefore, $\nu$ can be understood as the hyperparameter of the hyperparameters $\alpha,\beta$, which gives a hierarchical explanation of our model.
we give the concrete $q$-EM iteration for BLRS in Alg. \ref{alg:q-em-concrete}.
\begin{algorithm}[h]
   \caption{$q$-EM for BLRS}
% \begin{algorithmic}
\begin{enumerate}
\item Choose an initial setting for the parameters $\alpha,\beta$ as $\alpha^{\text{old}},\beta^{\text{old}}$;
%\item Estimation of $\nu$: optimizing $\nu$ by solving \eqref{eq:nu-estimate};
\item E Step: Evaluate $E_{p^{\prime}}(\w),E_{p^{\prime}}(\w\w^T)$ and $E_{p^{\prime}}((\w-\bmu)(\w-\bmu)^T)$ by \eqref{eq:A}, \eqref{eq:bmu}, \eqref{eq:B}, \eqref{eq:C} , \eqref{eq:epmu}, \eqref{eq:epsigma} with $\alpha=\alpha^{\text{old}},\beta=\beta^{\text{old}}$;
% \begin{eqnarray*} 
% E_{p^{\prime}}(\w)&=&(\beta^{\prime})^{\text{old}}\A^{-1}\bPhi^T\y,
% \end{eqnarray*}
% \begin{eqnarray*}
% &&E_{p^{\prime}}((\w-\bmu)(\w-\bmu)^T)\\
% \!\!\!\!&=&\!\!\!\!\!\frac{1}{\nu+m}(1+(\beta^{\prime})^{\text{old}}\y^T(\I-(\beta^{\prime})^{\text{old}}\bPhi \A^{-1}\bPhi^T)\y)A^{-1};
% \end{eqnarray*}
\item M Step: Evaluate $\alpha,\beta$ given by \eqref{eq:alpha-beta}.
% \[
% \btheta^{\text{new}} =\argmax_{\btheta} -D_q(p(\w|\y;\btheta^{\text{old}})\|p(\y,\w;\btheta));
% \] 
\item Check for convergence of either the log likelihood or the parameter values. If the convergence criterion is not satisfied, then let
\[
(\alpha^{\text{old}},\beta^{\text{old}})=(\alpha,\beta)
\]
and return to step 2.
\end{enumerate}
\label{alg:q-em-concrete}
% \end{algorithmic}
\end{algorithm}

In many regression tasks, the number of features are much less than the number of training samples, i.e., $M\ll m$. In order to reduce the computational complexity in each step as much as possible, we reformulate the steps in Alg. \ref{alg:q-em-concrete}. Before iteration, we give eigenvalues decomposition of $\bPhi^T\bPhi$ as 
\begin{eqnarray}
\bPhi^T\bPhi=\V\D\V^T,\label{eq:vdv}
\end{eqnarray}
where the time cost is $O(M^3)$ and compute
\begin{eqnarray}
\y_p=\bPhi^T\y, \quad \y_{pV}=\V^T\y_p, \quad \|\y\|_2^2,\label{eq:y-exp}
\end{eqnarray}
where the time costs are $O(mM)$, $O(M^2)$, and $O(m)$ respectively.
Then in each iteration,
\begin{eqnarray}
\bmu&=&(\bPhi^T\bPhi+\frac{\alpha}{\beta} \I)^{-1}\bPhi^T\y,\nonumber\\
&=&\V(\D+\frac{\alpha}{\beta}\I)^{-1}\y_{pV}, \label{eq:bmu-exp}
\end{eqnarray}
where the time cost is $O(M^2)$. Then 
\begin{eqnarray}
\y^T\B^{-1}\y&=&\beta\y^T(\I-\bPhi(\frac{\alpha}{\beta}\I+\bPhi^T\bPhi)^{-1}\bPhi^T)\y\nonumber\\
&=&\beta(\|\y\|_2^2-\y_p^T\bmu), \label{eq:yby}
\end{eqnarray}
where based on the result \eqref{eq:y-exp}, the time cost is $O(M)$. Then 
\begin{eqnarray}
\text{tr}(\C)&=&\frac{\nu+\y^T\B^{-1}\y}{\nu+m}\tr\left(\left(\alpha\I+\beta\bPhi^{T}\bPhi\right)^{-1}\right)\nonumber\\
&=&\frac{\nu+\y^T\B^{-1}\y}{\nu+m}\tr\left((\alpha\I+\beta\D)^{-1}\right), \label{eq:C-exp}
\end{eqnarray}
where based on the result \eqref{eq:yby}, the time cost is $O(M)$. Then
\begin{eqnarray}
&&\tr(\bPhi^T\bPhi\C)\nonumber\\
&=&\frac{\nu+\y^T\B^{-1}\y}{\nu+m}\tr(\V\D\V^T\V(\alpha\I+\beta\D)^{-1}\V^T)\nonumber\\
&=&\frac{\nu+\y^T\B^{-1}\y}{\nu+m}\tr(\D(\alpha\I+\beta\D)^{-1}),\label{eq:phi-C}
\end{eqnarray}
where based on the result \eqref{eq:yby}, the time cost is $O(M)$.  In addition
\begin{eqnarray}
\|\y-\bPhi\bmu\|_2^2=\|\y\|_2^2-2\y_p^T\bmu+\|\bPhi\bmu\|_2^2,\label{eq:y-bmu}
\end{eqnarray}
where based on the result \eqref{eq:y-exp}, the time cost is $O(mM+M)$.

Therefore, combing \eqref{eq:bmu-exp} and \eqref{eq:C-exp}, the time cost of computing $b$ in \eqref{eq:b} is $O(M^2+M)$; combing \eqref{eq:phi-C} and \eqref{eq:y-bmu}, the time cost of computing $c$ is $O(mM+M)$. For the BLRG case $\nu\rightarrow+\infty$,
$\frac{\nu+\y^T\B^{-1}\y}{\nu+m}\rightarrow1$, thus we can neglect the computation of $\y^T\B^{-1}\y$, which costs $O(M)$. However, in each iteration, the computational complexity is dominated by $O(mM+M^2)$, therefore compared with BLRG,  the extra cost of computing $\y^T\B^{-1}\y$ can be neglected. Based on this fact, when comparing the computational complexity between $q$-EM for BLRS and EM for BLRG in Section \ref{sec:exp}, we use the \emph{iteration count} as evaluation criterion.

% \begin{eqnarray}
% \A&=&\left(\alpha\I+\beta\bPhi^{T}\bPhi\right)^{-1},\label{eq:A}\\
% \bmu&=&\beta \A\bPhi^{T}\y=(\bPhi^T\bPhi+\frac{\alpha}{\beta} \I)^{-1}\bPhi^T\y.,\label{eq:bmu}\\
% \B&=&\beta^{-1}\I+\alpha^{-1}\bPhi\bPhi^T.\label{eq:B}
% \end{eqnarray}
\begin{section}{Experiments}\label{sec:exp}
We illustrate the performance of the $q$-EM algorithm for BLRS on the task of online news popularity prediction\footnote{The online news prediction dataset is available at \url{http://archive.ics.uci.edu/ml/datasets.html}.}, which is a trendy research topic \cite{fernandes2015proactive}.
 In experiments, the group of fixed basis functions $\{\phi_i(\cdot)\}_{i=1}^M$ is set to identity matrix. Each column of features is normalized with mean $0$ and unit length.
Then this dataset is divided into $5$ parts. In each trial, $4$ parts of them are used to learning the hyperparameters $\alpha,\beta$. Both the initial values of $\alpha,\beta$ are set to $1$.
We test $5$ values $\{10^{-8},10^{-5},10^{-2},10,10^4,+\infty\}$ of $\nu$ and list the maximum likelihood solution of $\alpha,\beta$ and the iteration count (cnt) with stopping criterion 
 \begin{eqnarray}
 \frac{|\alpha-\alpha^{\text{old}}|}{\alpha}<10^{-7}\quad\text{and}\quad \frac{|\beta-\beta^{\text{old}}|}{\beta}<10^{-7}
 \end{eqnarray}
on Table 1. Each subtable stands for the result on one of the $4$ divided parts of dataset. 

 Firstly it is shown that although both $q$-EM for BLRS and EM for BLRG are local minimizers, but they tend to find the same maximum likelihood solution of $\alpha,\beta$. Concretely, from each subtable of Table 1, all the $\alpha$'s are equal in the precision ($10^{-7}$) range and all the $\beta$'s are strictly equal in $10$ significant figures, which reflects the theoretical result in Theorem \ref{thm:1} to a certain extent.
 After $\alpha,\beta$ are determined, the mean $\bmu$ of $\w$ is determined uniquely. 
 Thus finding the same maximum likelihood solution of $\alpha,\beta$ means all the training error and test error will be the same. Therefore, in this experiments, we neglect the evaluation of performance and
 mainly concern the comparison of computational complexity between $q$-EM and EM.

Although $\nu$ is not related to the maximum likelihood solution of $\alpha,\beta$, we can see that the value of $\nu$ influences the convergence speed. Compared with $\nu\rightarrow+\infty$, by setting $\nu=10^{-8},10^{-5}$, $q$-EM can get $26.4\%,31.0\%,19.8\%,26.9\%,30.2\%$ speedup from the first trial to the fifth trial respectively in terms of iteration count. This is a remarkable result, because the iterate procedure of $q$-EM for BLRS is identical to EM for BLRG, except the coefficient $\frac{\nu+\y^T\B^{-1}\y}{\nu+m}$ of covariance matrix $\C$.

\end{section}

\begin{section}{Conclusions}
In this paper, we generalized the framework of Bayesian linear regression with Gaussian assumptions (BLRG) to Bayesian linear regression with Student-t assumptions (BLRS). Both the model and algorithm are generalized. We prove the equivalence of the maximum likelihood solution between the two models. Then we give the concrete iterative procedure of $q$-EM for BLRS, which is nearly identical to EM for BLRG. Finally experiments show the speedup effect by selecting the degrees of freedom $\nu$ to a positive constant.  
\vspace{0.3in}
\begin{table}[!th]
\caption{Maximum likelihood solutions of $\alpha,\beta$ for $5$ trials}
\begin{tabular}{|c|c|c|c|}
\hline
$\nu$	&	$\alpha$		&	$\beta$			&	cnt\\
\hline
\multicolumn{4}{|c|}{Trial 1}\\
\hline
1.0E-08	&	1.798414968E+12	&	1.426270263E+08	&	71\\ 
\hline
1.0E-05	&	1.798414964E+12	&	1.426270263E+08	&	71\\ 
\hline
1.0E-02	&	1.798414951E+12	&	1.426270263E+08	&	76\\ 
\hline
1.0E+01	&	1.798414896E+12	&	1.426270263E+08	&	85\\ 
\hline
1.0E+04	&	1.798414938E+12	&	1.426270263E+08	&	108\\ 
\hline
$+\infty$	&	1.798414906E+12	&	1.426270263E+08	&	110\\ 
\hline
% \end{tabular}
% \begin{tabular}{|c|c|c|c|}
\multicolumn{4}{|c|}{Trial 2}\\
% \hline
% $\nu$	&	$\alpha$		&	$\beta$			&	count\\
\hline
1.0E-08	&	2.815077496E+12	&	1.410306070E+08	&	69\\ 
\hline
1.0E-05	&	2.815077488E+12	&	1.410306070E+08	&	69\\ 
\hline
1.0E-02	&	2.815077620E+12	&	1.410306070E+08	&	75\\ 
\hline
1.0E+01	&	2.815077488E+12	&	1.410306070E+08	&	83\\ 
\hline
1.0E+04	&	2.815077464E+12	&	1.410306070E+08	&	97\\ 
\hline
$+\infty$	&	2.815077655E+12	&	1.410306070E+08	&	100\\ 
\hline

\multicolumn{4}{|c|}{Trial 3}\\

% \end{tabular}
% \begin{tabular}{|c|c|c|c|}

% \hline
% $\nu$	&	$\alpha$		&	$\beta$			&	cnt\\
\hline
1.0E-08	&	2.033476853E+12	&	1.767023424E+08	&	73\\ 
\hline
1.0E-05	&	2.033476847E+12	&	1.767023424E+08	&	73\\ 
\hline
1.0E-02	&	2.033476850E+12	&	1.767023424E+08	&	77\\ 
\hline
1.0E+01	&	2.033476856E+12	&	1.767023424E+08	&	82\\ 
\hline
1.0E+04	&	2.033476863E+12	&	1.767023424E+08	&	89\\ 
\hline
$+\infty$	&	2.033476869E+12	&	1.767023424E+08	&	91\\ 
\hline
\multicolumn{4}{|c|}{Trial 4}\\
\hline
1.0E-08	&	2.969154178E+12	&	1.162327242E+08	&	68\\ 
\hline
1.0E-05	&	2.969154172E+12	&	1.162327242E+08	&	68\\ 
\hline
1.0E-02	&	2.969154144E+12	&	1.162327242E+08	&	72\\ 
\hline
1.0E+01	&	2.969154128E+12	&	1.162327242E+08	&	78\\ 
\hline
1.0E+04	&	2.969154108E+12	&	1.162327242E+08	&	90\\ 
\hline
$+\infty$	&	2.969154291E+12	&	1.162327242E+08	&	93\\ 
\hline
\multicolumn{4}{|c|}{Trial 5}\\
\hline
1.0E-08	&	3.494880354E+12	&	1.209243318E+08	&	60\\ 
\hline
1.0E-05	&	3.494880348E+12	&	1.209243318E+08	&	60\\ 
\hline
1.0E-02	&	3.494880333E+12	&	1.209243318E+08	&	64\\ 
\hline
1.0E+01	&	3.494880373E+12	&	1.209243318E+08	&	71\\ 
\hline
1.0E+04	&	3.494880389E+12	&	1.209243318E+08	&	84\\ 
\hline
$+\infty$	&	3.494880355E+12	&	1.209243318E+08	&	86\\ 
\hline
\end{tabular}
 \end{table}		

 % hus finding the same maximum likelihood solution of $\alpha,\beta$ means all the training error and test error will be the same. Therefore, in this experiments, we mainly concern the comparison of computational complexity between $q$-EM and EM.

 % hus finding the same maximum likelihood solution of $\alpha,\beta$ means all the training error and test error will be the same. Therefore, in this experiments, we mainly concern the comparison of computational complexity between $q$-EM and EM.

 % hus finding the same maximum likelihood solution of $\alpha,\beta$ means all the training error and test error will be the same. Therefore, in this experiments, we mainly concern the comparison of computational complexity between $q$-EM and EM.

 % hus finding the same maximum likelihood solution of $\alpha,\beta$ means all the training error and test error will be the same. Therefore, in this experiments, we mainly concern the comparison of computational complexity between $q$-EM and EM.

\end{section}

\bibliographystyle{IEEEtran}
\bibliography{ML}

% Generated by IEEEtran.bst, version: 1.13 (2008/09/30)
\begin{thebibliography}{10}
\providecommand{\url}[1]{#1}
\csname url@samestyle\endcsname
\providecommand{\newblock}{\relax}
\providecommand{\bibinfo}[2]{#2}
\providecommand{\BIBentrySTDinterwordspacing}{\spaceskip=0pt\relax}
\providecommand{\BIBentryALTinterwordstretchfactor}{4}
\providecommand{\BIBentryALTinterwordspacing}{\spaceskip=\fontdimen2\font plus
\BIBentryALTinterwordstretchfactor\fontdimen3\font minus
  \fontdimen4\font\relax}
\providecommand{\BIBforeignlanguage}[2]{{%
\expandafter\ifx\csname l@#1\endcsname\relax
\typeout{** WARNING: IEEEtran.bst: No hyphenation pattern has been}%
\typeout{** loaded for the language `#1'. Using the pattern for}%
\typeout{** the default language instead.}%
\else
\language=\csname l@#1\endcsname
\fi
#2}}
\providecommand{\BIBdecl}{\relax}
\BIBdecl

\bibitem{shao1993linear}
J.~Shao, ``Linear model selection by cross-validation,'' \emph{Journal of the
  American statistical Association}, vol.~88, no. 422, pp. 486--494, 1993.

\bibitem{tsallis2001nonextensive}
C.~Tsallis, ``I. nonextensive statistical mechanics and thermodynamics:
  Historical background and present status,'' in \emph{Nonextensive statistical
  mechanics and its applications}.\hskip 1em plus 0.5em minus 0.4em\relax
  Springer, 2001, pp. 3--98.

\bibitem{de1997student}
A.~C. de~Souza and C.~Tsallis, ``Student's t-and r-distributions: Unified
  derivation from an entropic variational principle,'' \emph{Physica A:
  Statistical Mechanics and its Applications}, vol. 236, no.~1, pp. 52--57,
  1997.

\bibitem{matsuyama2003alpha}
Y.~Matsuyama, ``The $\alpha$-em algorithm: Surrogate likelihood maximization
  using $\alpha$-logarithmic information measures,'' \emph{Information Theory,
  IEEE Transactions on}, vol.~49, no.~3, pp. 692--706, 2003.

\bibitem{tsallis1988possible}
C.~Tsallis, ``Possible generalization of boltzmann-gibbs statistics,''
  \emph{Journal of statistical physics}, vol.~52, no. 1-2, pp. 479--487, 1988.

\bibitem{martins2009nonextensive}
A.~F. Martins, N.~A. Smith, E.~P. Xing, P.~M. Aguiar, and M.~A. Figueiredo,
  ``Nonextensive information theoretic kernels on measures,'' \emph{The Journal
  of Machine Learning Research}, vol.~10, pp. 935--975, 2009.

\bibitem{ding2010t}
N.~Ding and S.~Vishwanathan, ``t-logistic regression,'' in \emph{Advances in
  Neural Information Processing Systems}, 2010, pp. 514--522.

\bibitem{cichocki2010families}
A.~Cichocki and S.-i. Amari, ``Families of alpha-beta-and gamma-divergences:
  Flexible and robust measures of similarities,'' \emph{Entropy}, vol.~12,
  no.~6, pp. 1532--1568, 2010.

\bibitem{hoerl1970ridge}
A.~E. Hoerl and R.~W. Kennard, ``Ridge regression: Biased estimation for
  nonorthogonal problems,'' \emph{Technometrics}, vol.~12, no.~1, pp. 55--67,
  1970.

\bibitem{redner1984mixture}
R.~A. Redner and H.~F. Walker, ``Mixture densities, maximum likelihood and the
  em algorithm,'' \emph{SIAM review}, vol.~26, no.~2, pp. 195--239, 1984.

\bibitem{diaconis1979conjugate}
P.~Diaconis, D.~Ylvisaker \emph{et~al.}, ``Conjugate priors for exponential
  families,'' \emph{The Annals of statistics}, vol.~7, no.~2, pp. 269--281,
  1979.

\bibitem{tzikas2008variational}
D.~G. Tzikas, A.~C. Likas, and N.~P. Galatsanos, ``The variational
  approximation for bayesian inference,'' \emph{Signal Processing Magazine,
  IEEE}, vol.~25, no.~6, pp. 131--146, 2008.

\bibitem{rasmussen2004gaussian}
C.~E. Rasmussen, ``Gaussian processes in machine learning,'' in \emph{Advanced
  lectures on machine learning}.\hskip 1em plus 0.5em minus 0.4em\relax
  Springer, 2004, pp. 63--71.

\bibitem{bishop2006pattern}
C.~M. Bishop, \emph{Pattern recognition and machine learning}.\hskip 1em plus
  0.5em minus 0.4em\relax springer, 2006.

\bibitem{fernandes2015proactive}
K.~Fernandes, P.~Vinagre, and P.~Cortez, ``A proactive intelligent decision
  support system for predicting the popularity of online news,'' in
  \emph{Progress in Artificial Intelligence}.\hskip 1em plus 0.5em minus
  0.4em\relax Springer, 2015, pp. 535--546.

\end{thebibliography}
\end{document}